\documentclass[runningheads]{llncs}
\usepackage{graphicx}
\usepackage{amsmath,amssymb} 
\usepackage{color}
\usepackage{multirow}

\begin{document}
\pagestyle{headings}
\mainmatter

\def\ACCV20SubNumber{0000}  

\title{Gaussian Vector: An Efficient Solution for Facial Landmark Detection} 
\titlerunning{Gaussian Vector}
%
\author{Yilin Xiong\inst{1,2}\orcidID{0000-0003-1989-916X} \and
 Zijian Zhou\inst{2}\orcidID{0000-0003-3315-3962} \and
 Yuhao Dou\inst{2}\orcidID{0000-0002-4407-9609} \and
Zhizhong Su\inst{2}\orcidID{0000-0003-2312-9985}}
\authorrunning{Y.Xiong et al.}
%
\institute{Central South University, Changsha, Hunan, CN \and
Horizon Robotics \\ 
\email{yilin.xiong@csu.edu.cn} \\
\email{\{zijian.zhou, yuhao.dou, zhizhong.su\}@horizon.ai}}

\maketitle

\begin{abstract}
Significant progress has been made in facial landmark detection with the development of Convolutional Neural Networks.
The widely-used algorithms can be classified into coordinate regression methods and heatmap based methods.
However, the former loses spatial information, resulting in poor performance while the latter suffers from large output size or high post-processing complexity.
This paper proposes a new solution, Gaussian Vector, to preserve the spatial information as well as reduce the output size and simplify the post-processing.
Our method provides novel vector supervision and introduces Band Pooling Module to convert heatmap into a pair of vectors for each landmark.
This is a plug-and-play component which is simple and effective.
Moreover, Beyond Box Strategy is proposed to handle the landmarks out of the face bounding box.
We evaluate our method on 300W, COFW, WFLW and JD-landmark.
That the results significantly surpass previous works demonstrates the effectiveness of our approach.
\end{abstract}

\section{Introduction}

Facial landmark detection is a critical step for face-related computer vision applications, e.g. face recognition \cite{sun2014deep,liu2017sphereface,deng2019arcface}, face editing \cite{thies2016face2face} and face 3D reconstruction \cite{dou2017end,roth2015unconstrained,feng2018joint,zhu2017face}. Researchers have achieved great success in this field, especially after using Convolutional Neural Networks (CNNs). The popular algorithms can be divided into coordinate regression methods and heatmap based methods.

Coordinate regression methods use Fully Connected (FC) layers to predict facial landmark coordinates \cite{sun2013deep,zhang2014facial}.
However, the compacted feature maps before the final FC layers cause spatial information loss, leading to performance degradation.

\begin{figure}
	\centering
	\includegraphics[width=10.2cm]{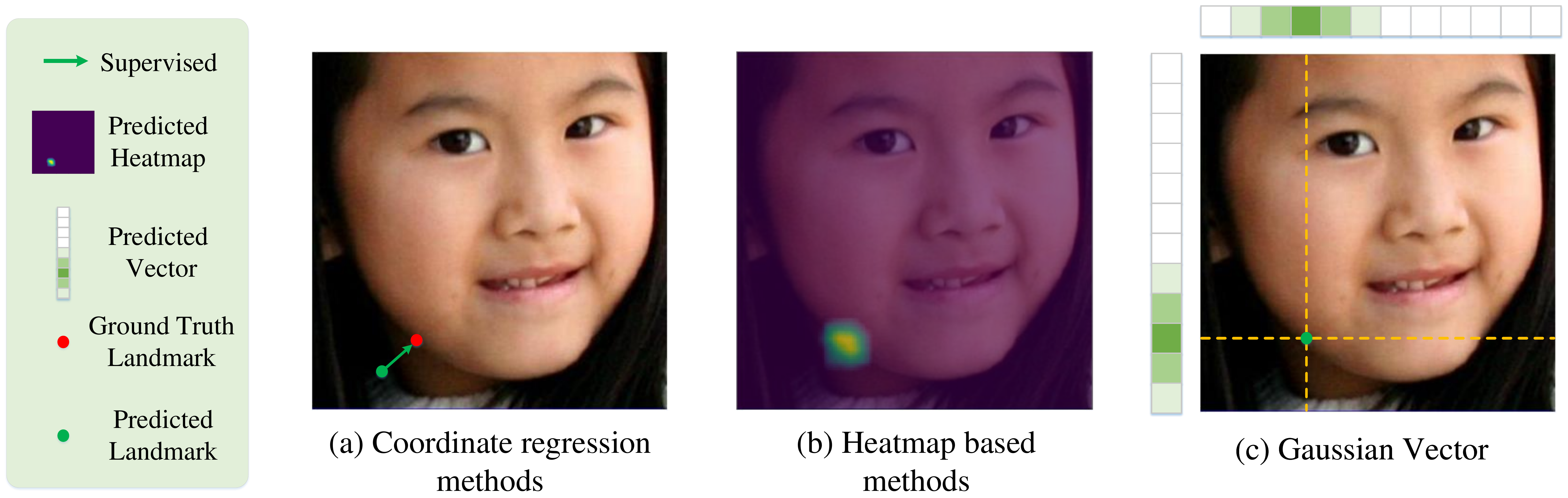}
	\caption{{\bf Comparison of existing methods and our proposed method.} 
		(a) coordinate regression methods and (b) heatmap based methods utilize the scalar coordinates and the heatmap to represent the spatial location of landmarks respectively.
		(c) Gaussian Vector locates a landmark via a pair of vectors.
	}
	\label{fig:diff_methods}
\end{figure}

As the advent of Fully Convolution Network \cite{long2015fully} and encoder-decoder structure \cite{ronneberger2015u}, a variety of heatmap based methods \cite{zhao2020mobilefan,wang2019adaptive,kowalski2017deep,wu2018look,newell2016stacked,dong2018supervision} sprung up to preserve the spatial information.
Heatmap based methods predict probability response maps, usually called heatmap, where each pixel predicts the probability that it is the landmark location, and find the highest response position by argmax operation. 
Heatmap based methods boost the performance compared with coordinate regression methods by better utilizing spatial information.

However, heatmap based methods suffer from following problems.
\textbf{1)} The large output heatmap tensor and complicated post-processing including argmax operation cause heavy burden of data transmission and computation in embedded systems. 
\textbf{2)} The ground truth heatmap consists of a small patch of foreground pixels with positive values and the remaining is full of background pixels with zero values. The background dominates the training process, which causes slow convergence. 
\textbf{3)} Only very few pixels around the highest response position is exploited to locate the landmark. 
Spatial information is not fully used, causing detection error. 
\textbf{4)} The input of landmark detection model is the face region of interest (RoI) accquried by a face detector. An imperfect face detector causes face bounding box shifting  resulting in some of the facial landmarks out of the bounding box. 
It is difficult for the heatmap based methods to construct a Gaussian surface to locate the landmarks out of the bounding box.

In this case, we are motivated to propose a new facial landmark detection algorithm, which not only maintains the spatial information but also overcome the problems above. 

\textbf{First of all}, inspired by the heatmap based methods, we use Gaussian formula to encode landmark coordinates into vectors as supervision. 
Vector label raises the proportion of foreground pixels in ground truth, which is beneficial to convergence. 
\textbf{Next}, Band Pooling Module (BPM) is proposed to convert the $h\times w$ output heatmap into $h\times 1$ and $1\times w$ vectors as prediction. 
On the one hand, vector prediction greatly eases the data transmission burden and simplifies the post-processing. 
On the other hand, pooling operation helps the network take advantage of more spatial information, which further improves the performance. 
\textbf{Last but not least}, we propose Beyond Box Strategy to handle landmarks out of the bounding box. 
The strategy helps our proposed method to reduce the error caused by bounding box shifting. 

We compare our approach with both coordinate regression methods and heatmap based methods in Fig.~\ref{fig:diff_methods}. Considering the utilization of Gaussian formula and vector prediction, we call our method Gaussian Vector (GV).

We evaluate our method on four popular facial landmark detection benchmarks including 300W \cite{sagonas2013300}, COFW \cite{burgos2013robust}, WFLW \cite{wu2018look} and the recent Grand Challenge of 106-p Facial Landmark Localization benchmark \cite{liu2019grand} (JD-landmark for short). 
Our approach outperforms state-of-the-art methods in 300W, COFW and WFLW. 
Without bells and whistles, we achieve the second place in JD-landmark by utilizing a single model.

In summary, our main contributions include:

\begin{itemize}
	\item We propose an efficient method, Gaussian Vector, for facial landmark detection, which boosts the model performance and reduces the system complexity.
	\item We introduce novel vector label to encode landmark location for supervision, which alleviates the imbalance of foreground and background in heatmap supervision.
	\item We design a plug-and-play module, Band Pooling Module (BPM), that effectively converts heatmap into vectors. 
	By this means, the model takes more spatial information into account yet outputs smaller tensor.
	\item We propose the Beyond Box Strategy, that enables our method to predict landmarks out of the face bounding box.
	\item Our method achieves state-of-the-art results on 300W, COFW, WFLW and JD-landmark.
\end{itemize}

\section{Related Work}

Facial landmark detection, or face alignment, has become a hot topic in computer vision for decades. 
Early classical methods include Active Appearance Models (AAMs) \cite{cootes2001active,kahraman2007active,matthews2004active}, Active Shape Models (ASMs) \cite{milborrow2008locating}, Constrained Local Models (CLMs) \cite{cristinacce2006feature}, Cascaded Regression Models \cite{cao2014face,burgos2013robust,chen2014joint,xiong2015global} and so on. 
With the development of deep learning, CNN-based methods have dominated this field.
The prevailing approaches fall into two categories.

{\bf Coordinate regression methods} take the face image as input and output landmark coordinates directly. This kind of methods get better results than the classic without extra post-processing. 
Sun \textit{et al.} \cite{sun2013deep} borrowed the idea of Cascaded Regression and combined serial sub-networks to reduce the location error. 
TCDCN \cite{zhang2014facial} researched the possibility of improving detection robustness through multi-task learning. 
MDM \cite{trigeorgis2016mnemonic}  employed Recurrent Neural Network (RNN) to replace the cascaded structure in Cascaded Regression. 
Feng \textit{et al.} \cite{feng2018wing}  presented a new loss function to pay more attention to small and medium error. 
Most of the approaches above are proposed with cascaded structure, which largely increases the system complexity.

{\bf Heatmap based methods} are popular in recent years. 
Benefit from the encoder-decoder structure, heatmap based methods preserve spatial information by upsampling and significantly improve state-of-the-art performance on all benchmarks.
DAN \cite{kowalski2017deep} utilized the heatmap information for each stage in the cascaded network and reduced the localization error. 
Stacked Hourglass \cite{newell2016stacked} introduces repeated bottom-up, top-down architecture for points detection and proposes a shift-based post-processing strategy.
HRNet \cite{sun2019high,sun2019deep} presented a novel network architecture for landmark detection, which maintained high-resolution representations through the design of dense fusions on cross-stride layers.
LAB \cite{wu2018look} proposed a boundary-aware method, which used the stacked hourglass \cite{newell2016stacked} to regress the boundary heatmap for capturing the face geometry structure. 
Another landmarks regressor fused the sturcture information in multiple stages and got the final results.
AWing \cite{wang2019adaptive} followed the work of Wing loss \cite{feng2018wing} and employed the revised version in the heatmap based methods.
Also, boundary information and CoordConv \cite{liu2018intriguing} are utilized for promoting landmark detection. 

However, recovering landmark coordinates from the heatmap still remains to be simplified. 
Efforts has been devoted to improve the efficiency of this process recently \cite{sun2018integral,nibali2018numerical,levine2016end}. 
Specifically, Sun \textit{et al.} \cite{sun2018integral} proposed a differentiable operation to convert heatmap into numerical coordinates for end-to-end training. 
Yet the calculation of exponent is not computation-efficient in embedded systems.

We believe that brand new supervision is necessary to address problems above. 
To this end, we encode the location information into vectors for supervision. 

\section{Our approach}\label{section:approach}
We first present an overview of our approach in Sec.~\ref{section:overall}. 
Then Sec.~\ref{section:label} introduces the generation of vector label.
After that, we illustrate the implementation details of our core component BPM in Sec.~\ref{section:BPM}.
Finally, we propose a strategy to predict the landmarks out of the face bounding box in Sec.~\ref{section:beyond}.
\subsection{Overview}\label{section:overall}
\begin{figure}
	\centering
	\includegraphics[width=10.5cm]{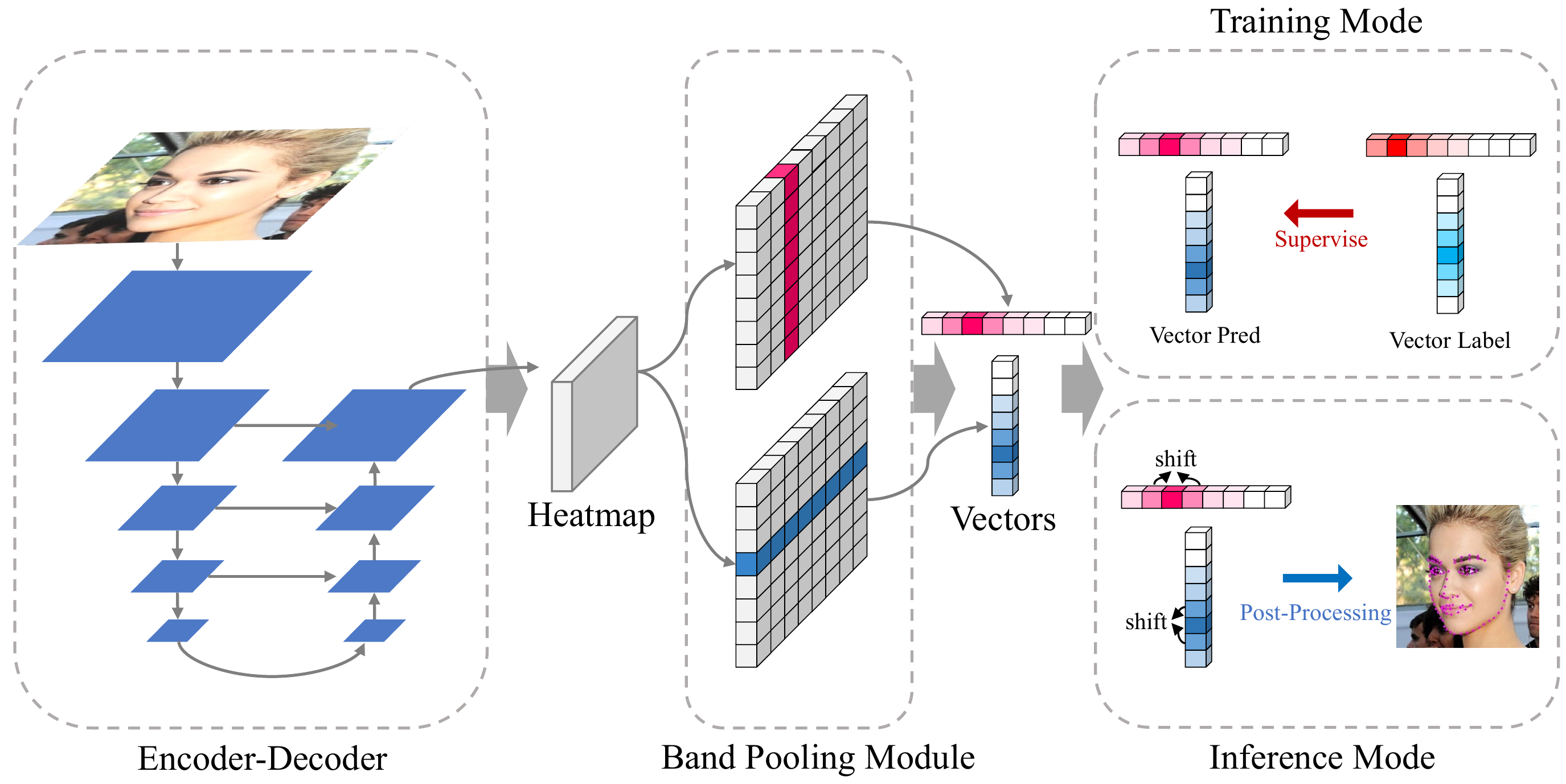}
	\caption{\textbf{The overview of the proposed method.}
		Given the face image inputs to the encoder-decoder structure, we can obtain the output heatmap and generate the prediction vectors through Band Pooling Module, match it with vector label for training.
		Landmark coordinates can be recovered from the vectors by locating the maximum value indexes.
		Shift strategy \cite{newell2016stacked} is used to reduce  quantization error.
	}
	\label{fig:arch}
\end{figure}

An overview of our pipeline is shown in Fig.~\ref{fig:arch}.
Similar to heatmap based methods, we use the encoder-decoder structure in our model. 
The encoder, with fully convolution layers, extracts spatial information on different scales.
For the decoder, Feature Pyramid Network (FPN) \cite{lin2017feature} is adopted.
It combines the features of both higher and lower levels through the top-down and botton-up pathways, which improves the pixel-level detection tasks.

The network input is cropped and normalized face image.
The network outputs 1/4 resolution heatmap to the original input, where each channel is corresponding to one landmark. 
A plug-and-play component Band Pooling Module (BPM) compresses heatmap into vectors for end-to-end training.

During training, the predicted vector is supervised by the vector label which is generated from Gaussian formula. 
More details about vector label can be found in Sec.~\ref{section:label}. 
Mean Square Error (MSE) is used as loss function.

During inference, the maximum of predicted vectors indicates the horizontal and vertical location of the landmark respectively. 
A shift strategy \cite{newell2016stacked}, widely used in heatmap based methods, is applied according to its neighbor pixels to reduce location error.

\subsection{Vector Label}\label{section:label}
Vector label $\hat{\bf G}=(\hat{\bf x},\hat{\bf y})$ for each landmark is generated by the Gaussian formula. 
Take vector $\hat{\bf x}\in \mathbb{R}^w$ as an example, where $w$ means the width of the output heatmap.
For the landmark coordinate $p(x_{0},y_{0})$, we first calculate the Euclidean distance between each pixel and $x_{0}$ in the vector, getting the distance vector ${\bf d}=\begin{Bmatrix}d_{x}^{i}\end{Bmatrix}_{i=0}^{w-1}$, each $d_{x}^{i}$ as:
\begin{equation}
d_{x}^{i}=\sqrt{(x_{i}-x_{0})^2}, i\in\left [0,w\right )
\end{equation} 

Then, we use quasi-Gaussian distribution to transform the distance vector $\bf d$ to vector label $\hat{\bf x}=\left \{{\hat x}^{i} \right \}^{w-1}_{i=0}$ with standard deviation $\sigma$. The elements are defined as:
\begin{equation}
\label{equ.supervised}
\hat{x}^{i}=
\begin{cases}
e^{-\frac{{d_{x}^{i}}^{2}}{2\sigma ^{2}}},  & \text{if $d_{x}^{i}<3\sigma$ and $d_{x}^{i}\neq0$} \\
1+\theta,  & \text{if $d_{x}^{i}=0$} \\
0, & \text{otherwise}
\end{cases}
\end{equation}

We recommend $\sigma$=2 in most cases.
Compared with the normal Gaussian distribution, we reinforce the distribution peak to enhance the supervision by adding a positive constant $\theta$.
By this enhancement, we enlarge the value gaps between the horizontal landmark location $x_{0}$ and the others to avoid deviation of the highest response position.
In the same way, we can get ${\bf \hat{y}}\in \mathbb{R}^h$, where $h$ is the heatmap height. 

The proposed vector label is beneficial to the model training in two aspects.
On the one hand, generating vectors instead of heatmap accelerates the label preparing.
On the other hand, extreme foreground background imbalance, as mentioned above, is alleviated in a straightforward way.

\subsection{Band Pooling Module}\label{section:BPM}
\begin{figure}
	\centering
	\includegraphics[width=9.0cm]{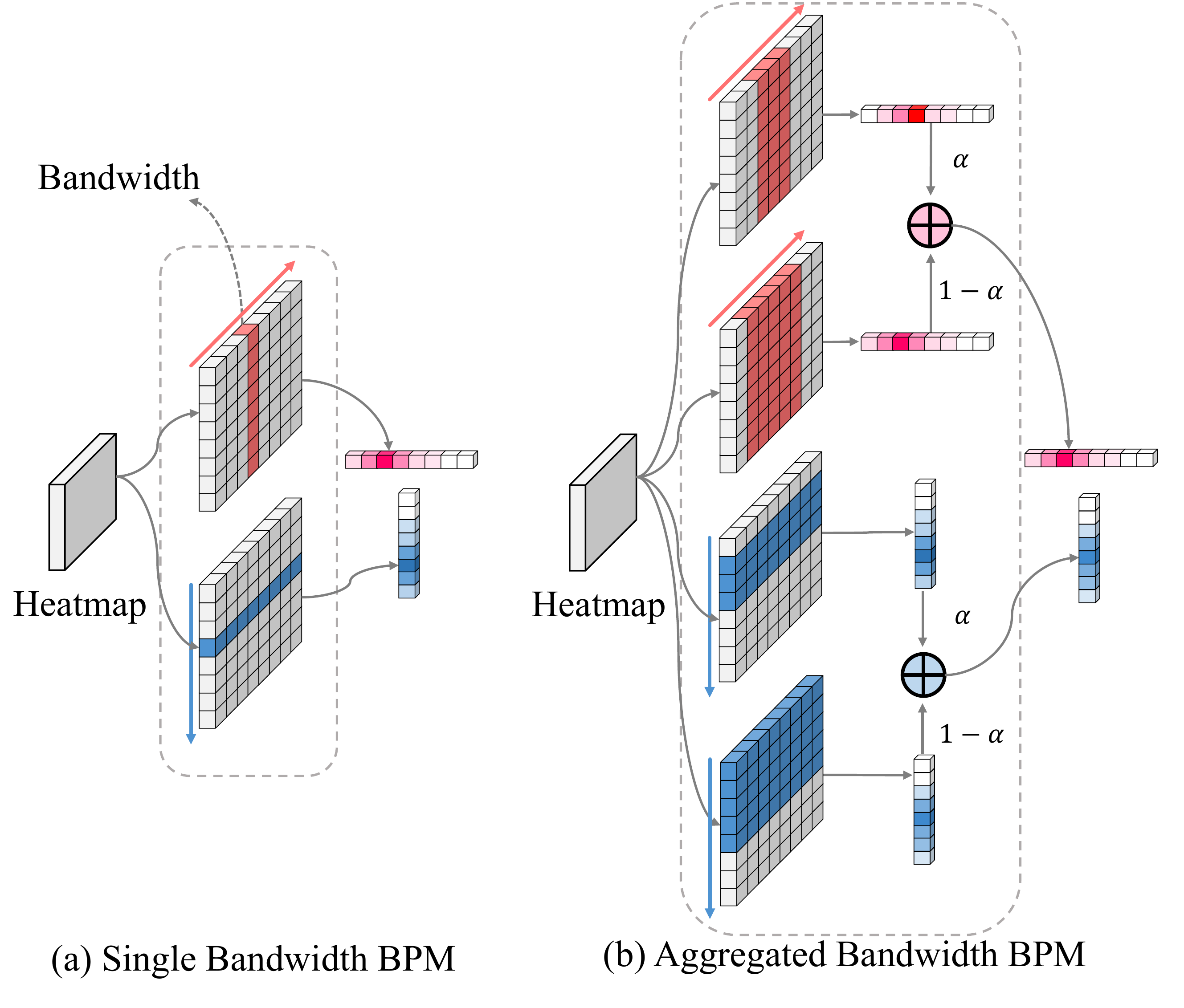}
	\caption{{\bf The illustration of BPM}.
		The vertical band $B_{v}$ (in \textit{red}) and the horizontal band $B_{h}$ (in \textit{blue}) slide on the output heatmap.
		We illustrate the vector generated by single bandwidth (a) and aggregated bandwidth (b).
		Specifically, the feature vectors of bandwidth $l$=3 and $l$=5 are able to aggregated by weighted elementwise sum.
	}
	\label{fig:BPM}
\end{figure}
For end-to-end training, a plug-and-play component Band Pooling Module is presented.
The BPM aims to perform the following conversion:
\begin{equation}
{\bf H}\rightarrow {\bf G}
\end{equation}
$\bf H$ is the output feature map and ${\bf G}=({\bf x},{\bf y})$ refers to the predicted vectors, matching with vector label $\hat{\bf G}$ for training.

Band Pooling Module, consisting of horizontal band $B_{h}\in \mathbb{R}^{l\times w}$ and vertical band $B_{v}\in \mathbb{R}^{h\times l}$, plays a crucial role in the process of generating predicted vectors.
The long side of the band is the same as side length of heatmap ($w$ for $B_{h}$ and $h$ for $B_{v}$).
The bandwidth, denoted as $l$, is a much smaller odd number.
The two long and thin bands slide on the output heatmap and average the values of pixels that fall into the band regions, generating predicted vectors ${\bf x}$ and ${\bf y}$. The pooling operation is shown in Fig.~\ref{fig:BPM}(a). It can be described as:

\begin{equation}
{\bf x}_{l}=AvgPooling(\textbf{H};B_{v},l)
\end{equation}

The bandwidth $l$ is adjustable, which controls the receptive field size of the vector elements.
Undersize bandwidth strictly limits the receptive field, resulting in loss of spatial information, which is essential for the landmark detection.
Conversely, overextended receptive field brings in redundant information from the background or nearby landmarks, leading to confusion.
In practical, bandwidth $l$ is suggested to be chosen from 1 and 7 and we show the effect of different $l$ in Table.~\ref{table:bandwidth}.

In addition, we extend the basic BPM through fusing the vectors generated by  different bandwidths.
In general considering the performance and complexity, we aggregate the vectors from only two different bandwidth $l_{1}$ and $l_{2}$ by weights.

\begin{equation}\label{equ:alpha}
{\bf x}=\alpha {\bf x}_{l_{1}}+(1-\alpha){\bf x}_{l_{2}}, 0<\alpha<1
\end{equation}
Generally, $\alpha$ is set to be 0.5.
Similarly, we can get ${\bf y}$. Zero padding is necessary on the edges when $l>$1. The aggregation is shown in Fig.~\ref{fig:BPM}(b).

In this way, the model comprehensively takes spatial information into account via the bigger bandwidth and pay more attention to the adjacent region through the smaller.

By using BPM, we narrow the search regions from 2D maps to 1D vectors when seeking the maximum of the prediction, leading to large reduction on post-processing complexity (from $O(N^{2})$ to $O(N)$).

\subsection{Beyond the Box}\label{section:beyond}
The face bounding box has a significant impact on facial landmark detection. 
An oversize bounding box brings in redundant information, especially in the face-crowded image. 
But a tight box may cause facial landmarks to locate out of the face bounding box. 
To solve this problem, we propose an effective strategy to predict outer landmarks.

\begin{figure}
	\centering
	\includegraphics[width=9.0cm]{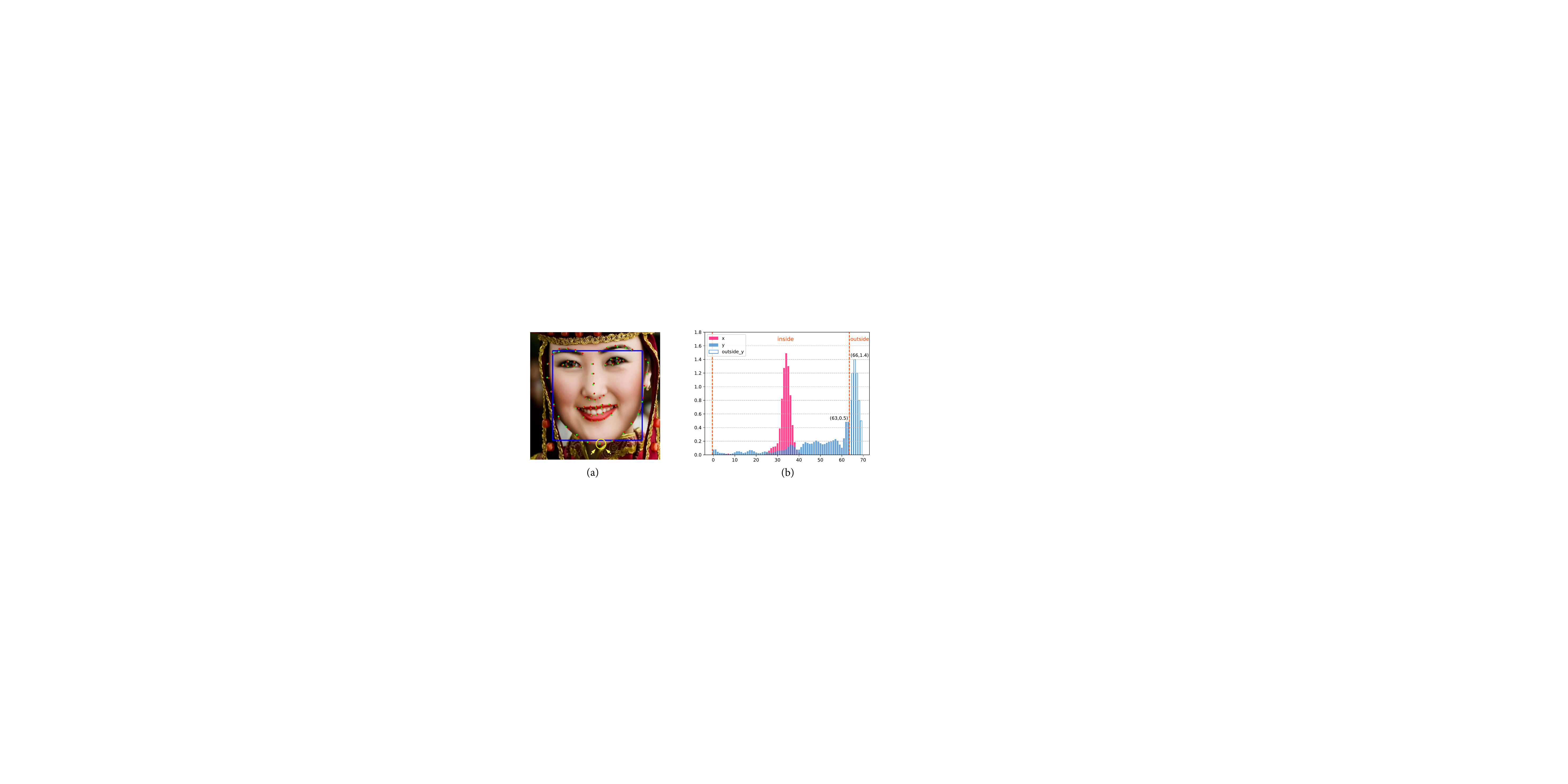}
	\caption{
		(a) \textbf{A visualization of the prediction with the Beyond Box Strategy}.
		The ground truth and prediction landmarks are shown in \textit{red} and \textit{green}, respectively.
		The \textit{blue} rectangle refers to the face bounding box.
		(b) 
		\textbf{We use a histogr7am to visualize the predicted vector of landmark in the \textit{yellow circle} of the \textit{left} image.}
		The \textit{red dashed line} shows the distribution of predicted vector inside, while the outside shows the estimated distribution from Beyond Box Strategy when the maximum locates at the endpoint of vector (e.g. $\bf y$).
		We calculate the position of ideal distribution peak, i.e. (66, 1.4) in histogram, according to Eq.~\ref{equ:dis}.
	}
	\label{fig:beyond}
\end{figure}

In general, the predicted vectors present smoother quasi-Gaussian distribution than vector label, where the maximum implies the landmark location. 
When the landmark locates inside, the maximum position is in the middle of the vector.
On the contray, when the landmark falls out of the input, the maximum is close to one of the endpoints and the quasi-Gaussian distribution is cut off.  
An example of the outside landmark is displayed in Fig.~\ref{fig:beyond}. 
Apparently, the maximum value of $\bf y$ is much smaller than the normal ones.

Fortunately, the maximum value tells us whether the landmark goes beyond and how far it goes.
If the distribution is cut off and the maximum locate at endpoint (0 or $w-1$ for index $s$), we assume $x^s$ obeys the distribution:
\begin{equation}
\label{equ:dis}
{x}^{s}=\tau e^{-\frac{d^{2}}{2\sigma^{2}}}
\end{equation}
where $d$ indicates the distance between $s$ and the distribution peak.
$\tau$ is a scale parameter and $\sigma$ is the standard deviation, which control the peak value and smoothness of distribution respectively.
The assumption helps the model to estimate the location of the distribution peak, which works only if the vector maximum locates at one of the endpoints, just like the blue vector in Fig.~\ref{fig:beyond}(b).

Assume that the maximum locates at the endpoint, we can get the revised landmark location:
\begin{equation}
\label{equ:loc}
loc=
\begin{cases}
-\sqrt{-2\sigma^{2}log(\frac{x^{0}}{\tau})} & \text{if $\mathop{argmax}\limits_{i}$ $x^{i}=0$} \\
w-1+\sqrt{-2\sigma^{2}log(\frac{x^{w-1}}{\tau})} & \text{if $\mathop{argmax}\limits_{i}$ $x^{i}=w-1$ }
\end{cases}
\end{equation}

Generally, $\tau$ ranges from $0.5\times(1+\theta)$ to $0.8\times(1+\theta)$ and $\sigma$ equals to that in Eq.~\ref{equ.supervised}.

\section{Experiments}
To verify the effectiveness of our method, a series of experiments are conducted on public datasets in Sec.~\ref{sec:eval}. 
Ablation study is carried out on 300W in Sec.~\ref{section:ablation} to test the proposed components.
In Sec.~\ref{sec:analysis}, we perform efficiency analysis to validate the points made above.
In addition, we give comprehensive analysis of the merits of vector supervision  in the supplement material.
\subsection{Datasets}
{\bf 300W} \cite{sagonas2013300} is a widely-used benchmark for facial landmark detection, which is a compound dataset including HELEN \cite{le2012interactive}, LFPW \cite{belhumeur2013localizing}, AFW \cite{zhu2012face}, and IBUG \cite{sagonas2013300} and re-annotated by the 68-point scheme. 
Same as the previous works \cite{wu2018look,zhao2020mobilefan}, we utilize 3148 images for training and 689 images for validation. 
The full test set can be divided into the common subset and the challenge subset.

{\bf COFW} \cite{burgos2013robust} is designed to present faces in realistic conditions. 
Faces in the pictures show large variations in shape and occlusion due to differences in expression, as well as pose. 
COFW has 1354 images for training and 507 images for testing. 
Each image is annotated with 29 points consistent with LFPW.

{\bf WFLW} \cite{wu2018look} is a recently proposed large dataset which consists of 7500 training images and 2500 testing images with 98-point annotation. 
It is one of the most challenging benchmarks for facial landmark detection because of existing extreme disturbance such as occlusion, large pose, illumination, and blur. 
In order to evaluate the model in variant conditions, the test dataset is categorized as multiple subsets.

{\bf JD-landmark} \cite{liu2019grand} is introduced for competition in 2019. 
Each image is manually annotated with 106-point landmarks. 
It contains three parts of training, validation, and test sets. 
The training set provides 11393 images with different poses. 
The validation and test sets come from web open datasets and cover large variations in pose, expression and occlusion, including 2000 images respectively.

\subsection{Evaluation Metrics}\label{section:metric}
To facilitate comparison with previous works, we adopt different evaluation metrics for different benchmarks including Normalized Mean Error (NME), Area Under Curve (AUC) based on Cumulative Error Distribution (CED) curve, and Failure Rate (FR).

NME is widely used in facial landmark detection to evaluate the quality of models. 
It is calculated by average the Euclidean distance between predicted and ground truth landmarks, then normalized to eliminate the impact caused by the image size inconsistency. 
NME for each image is defined as:
\begin{equation}
NME=\frac{1}{L}\sum^{L}_{k=1}\frac{\left \|p_{k}-\hat{p}_{k}\right \|_{2}}{d}
\end{equation}
Where $L$ refers to the number of landmarks. 
$p_{k}$ and $\hat{p}_{k}$ refer to the predicted and ground truth coordinates of the $k$-th landmark respectively. 
$d$ is the normalization factor, such as the distance of eye centers (inter-pupil normalization, IPN) or the distance of eye corners (inter-ocular normalization, ION). 
Specially, For JD-landmark, $d=\sqrt{w_{bbox}\times h_{bbox}}$, where $w_{bbox}$ and $h_{bbox}$ are the width and height of the enclosing rectangle of the ground truth landmarks.

FR is another widely-used metric.
It is the percentage of failure samples whose NME is larger than a threshold. 
As the threshold rises from zero to the target value, we plot the proportion of success (samples with smaller NME than the threshold) as the CED curve. 
AUC calculates the area under the CED curve.

\subsection{Implementation Details}
Firstly, for each face bounding box, we extend the short side to the same as the long  to avoid \textit{image distortion} since the bounding box is not square in most cases.
Then we enlarge the bounding boxes in WFLW and COFW by 25\% and 10\% respectively, and keep the extended square box size for 300W and JD-landmark.
Finally, the images are cropped and resized to $256\times 256$ according to the square boxes from the previous step.

In terms of data augmentation, we use standard methods: $30^{\circ}$ max angle of random rotation with the probability of 50\%, random scale 75\%$\sim$125\% for all images, random flip horizontally with the probability of 50\%.
We also try random occlusion in training which improves the performance. 
We optimize models with Adam \cite{kingmaadam} for all benchmarks. 
For bandwdith $l$, we choose both 3 and 5 for aggregating. 
For $\sigma$ in Eq.~\ref{equ.supervised}, it is set to 2 on all benchmarks.
Our models are trained with MXNet/Gluon 1.5.1 \cite{chen2015mxnet} on $4\times$ TITAN X GPUs.
During inference our baseline model costs 28ms per face for the entire pipeline (network forward and post-processing) with batch size = 1 and $1\times$ TITAN X.

\begin{minipage}{\textwidth}
	\begin{minipage}[t]{0.5\textwidth}
		\scriptsize
		\centering
		\renewcommand\arraystretch{1.4}
		\makeatletter\def\@captype{table}\makeatother
		\caption{NME (\%) on 300W Common, Challenging, Full subset}
		\label{table:300w}
		\begin{tabular}{ccccc}
			\cline{1-4}
			\multicolumn{1}{c}{Method} & \begin{tabular}[c]{@{}c@{}}Common\end{tabular} 
			& 
			\begin{tabular}[c]{@{}c@{}}Challenging\end{tabular} 
			& \multicolumn{1}{c}{Fullset} &  \\ \cline{1-4}
			\multicolumn{4}{c}{Inter-pupil Normalization}      &  \\ \cline{1-4}
			\multicolumn{1}{c|}{RCPR\cite{burgos2013robust}} & 6.18 & 17.26 & \multicolumn{1}{c}{8.35} &  \\
			\multicolumn{1}{c|}{LBF\cite{ren2014face}} & 4.95 & 11.98 & \multicolumn{1}{c}{6.32} &  \\
			\multicolumn{1}{c|}{TCDCN\cite{zhang2014facial}} & 4.80 & 6.80 & \multicolumn{1}{c}{5.54} &  \\
			\multicolumn{1}{c|}{RAR\cite{xiao2016robust}} & 4.12 & 8.35 & \multicolumn{1}{c}{4.94} &  \\
			\multicolumn{1}{c|}{LAB\cite{wu2018look}} & 4.20 & 7.41 & \multicolumn{1}{c}{4.92} &  \\
			\multicolumn{1}{c|}{DCFE\cite{valle2018deeply}} & 3.83 & 7.54 & \multicolumn{1}{c}{4.55} &  \\
			\multicolumn{1}{c|}{AWing\cite{wang2019adaptive}} & 3.77 & 6.52 & \multicolumn{1}{c}{4.31} &  \\ \cline{1-4}
			\multicolumn{1}{c|}{\bf GV(ResNet50)} & 3.78 & 6.98 & \multicolumn{1}{c}{4.41} & \\ 
			\multicolumn{1}{c|}{\bf GV(HRNet)}  & {\bf 3.63}   & {\bf 6.51}   & \multicolumn{1}{c}{\bf 4.19} &  \\
			\cline{1-4}
			\multicolumn{4}{c}{Inter-ocular Normalization}                              &  \\ \cline{1-4}
			\multicolumn{1}{c|}{DAN\cite{kowalski2017deep}} & 3.19 & 5.24 & \multicolumn{1}{c}{3.59} &  \\
			\multicolumn{1}{c|}{DCFE\cite{valle2018deeply}} & 2.76 & 5.22 & \multicolumn{1}{c}{3.24} &  \\
			\multicolumn{1}{c|}{LAB\cite{wu2018look}} & 2.98 & 5.19 & \multicolumn{1}{c}{3.49} &  \\
			\multicolumn{1}{c|}{HRNetV2\cite{sun2019high}} & 2.87 & 5.15 & \multicolumn{1}{c}{3.32} &  \\
			\multicolumn{1}{c|}{AWing\cite{wang2019adaptive}} & 2.72 & 4.52 & \multicolumn{1}{c}{3.07} &  \\ \cline{1-4}
			\multicolumn{1}{c|}{\bf GV(ResNet50)} & 2.73 & 4.84 & \multicolumn{1}{c}{3.14} &  \\ 
			\multicolumn{1}{c|}{\bf GV(HRNet)}   & {\bf 2.62} &  {\bf 4.51} & \multicolumn{1}{c}{\bf 2.99} &  \\
			\cline{1-4}
		\end{tabular}
	\end{minipage}
	\begin{minipage}[t]{0.5\textwidth}
		\scriptsize
		\centering
		\renewcommand\arraystretch{1.473}
		\makeatletter\def\@captype{table}\makeatother
		\setlength{\tabcolsep}{6pt}
		\caption{NME (\%) and FR (\%) on COFW testset}
		\label{table:cofw}
		\begin{tabular}{c|cc}
			\hline
			\multicolumn{1}{c}{Method}	& NME & $FR_{10\%}$ \\ \hline
			\multicolumn{3}{c}{Inter-pupil Normalization}   \\ \hline
			Human\cite{burgos2013robust}	& 5.60  & - \\
			RCPR\cite{burgos2013robust}	& 8.50 & 20.0 \\
			RAR\cite{xiao2016robust} & 6.03 & 4.14 \\ 
			TCDCN\cite{zhang2014facial}	& 8.05 & 20.0 \\
			DAC-CSR\cite{feng2017dynamic}	& 6.03 & 4.73 \\
			PCD-CNN\cite{kumar2018disentangling}	& 5.77 & 3.73 \\
			DCFE\cite{valle2018deeply} & 5.27 & - \\ 
			AWing\cite{wang2019adaptive}& 4.94 & 0.99 \\ \hline
			{\bf GV(ResNet50)} & 4.92 &  0.99 \\ 
			{\bf GV(HRNet)}	& {\bf 4.85} & {\bf 0.59} \\ \hline
			\multicolumn{3}{c}{Inter-ocular Normalization}   \\ \hline
			LAB\cite{feng2017dynamic}	& 3.92 & 0.39 \\
			MobileFAN\cite{zhao2020mobilefan}	& 3.66 & 0.59 \\
			HRNetV2\cite{sun2019high} & 3.45 & \textbf{0.19} \\ \hline
			\textbf{GV(ResNet50)} & 3.42& 0.59 \\
			\textbf{GV(HRNet)} & \textbf{3.37}& 0.39 \\ \hline
		\end{tabular}
	\end{minipage}
\end{minipage}

\subsection{Evaluation on different benchmarks} \label{sec:eval}
We provide two models based on ReNetV2-50 \cite{he2016identity} and HRNetV2-W32 \cite{sun2019high}, represented as GV(ResNet50) and GV(HRNet) respectively, to verify the generality and effectiveness of our approach.
Experiments show that our approach achieves state of the art on all benchmarks.\footnote{In addition, we analyze the merits of vector supervision in supplemental material.}

{\bf Evaluation on 300W}.
Our proposed method achieves the best performance on 300W as shown in Table~\ref{table:300w}. 
GV(ResNet50) achieves 3.14\% for ION (4.41\% for IPN), and GV(HRNet) achieves 2.99\% for ION (4.19\% for IPN). Both models get remarkable results.

\begin{table}[t]
	\addtolength{\tabcolsep}{-0.5pt}
	\scriptsize
	\renewcommand\arraystretch{1.2}
	\centering
	\caption{Evaluation on WFLW}
	\label{table:wflw}
	\begin{tabular}{ccccccccc}
		\hline
		Metric	&     Method     &     Testset      & \begin{tabular}[c]{@{}c@{}}Pose\\Subset\end{tabular}                     &          \begin{tabular}[c]{@{}c@{}}Expression\\Subset\end{tabular}            &           \begin{tabular}[c]{@{}c@{}}Illumination\\Subset\end{tabular}           &            \begin{tabular}[c]{@{}c@{}}Make-up\\Subset\end{tabular}          &     \begin{tabular}[c]{@{}c@{}}Occlusion\\Subset\end{tabular}                 &       \begin{tabular}[c]{@{}c@{}}Blur\\Subset\end{tabular}                \\ \hline
		\multirow{7}{*}{\begin{tabular}[c]{@{}c@{}}NME(\%)\\ (Lower is better)\end{tabular}} &    ESR\cite{cao2014face}               &       11.13               &            25.88          &       11.47               &    10.49                  &           11.05           &       13.75               &    12.20                  \\
		& \multicolumn{1}{c}{SDM\cite{xiong2013supervised}} & \multicolumn{1}{c}{10.29} & \multicolumn{1}{c}{24.10} &           \multicolumn{1}{c}{11.45} & \multicolumn{1}{c}{9.32} & \multicolumn{1}{c}{9.38} & \multicolumn{1}{c}{13.03} & \multicolumn{1}{c}{11.28} \\
		& \multicolumn{1}{c}{CFSS\cite{zhu2015face}} & \multicolumn{1}{c}{9.07} & \multicolumn{1}{c}{21.36} &           \multicolumn{1}{c}{10.09} & \multicolumn{1}{c}{8.30} & \multicolumn{1}{c}{8.74} & \multicolumn{1}{c}{11.76} & \multicolumn{1}{c}{9.96} \\
		& \multicolumn{1}{c}{DVLN\cite{wu2017leveraging}} & \multicolumn{1}{c}{6.08} & \multicolumn{1}{c}{11.54} &           \multicolumn{1}{c}{6.78} & \multicolumn{1}{c}{5.73} & \multicolumn{1}{c}{5.98} & \multicolumn{1}{c}{7.33} & \multicolumn{1}{c}{6.88} \\	
		&\multicolumn{1}{c}{LAB\cite{wu2018look}} & \multicolumn{1}{c}{5.27} & \multicolumn{1}{c}{10.24} & \multicolumn{1}{c}{5.51} &           \multicolumn{1}{c}{5.23} & \multicolumn{1}{c}{5.15} & \multicolumn{1}{c}{6.79} & \multicolumn{1}{c}{6.32}   \\
		& \multicolumn{1}{c}{Wing\cite{feng2018wing}} & \multicolumn{1}{c}{5.11} & \multicolumn{1}{c}{8.75} &           \multicolumn{1}{c}{5.36} & \multicolumn{1}{c}{4.93} & \multicolumn{1}{c}{5.41} & \multicolumn{1}{c}{6.37} & \multicolumn{1}{c}{5.81} \\
		& \multicolumn{1}{c}{AWing\cite{wang2019adaptive}} & \multicolumn{1}{c}{4.36} & \multicolumn{1}{c}{\bf 7.38} &           \multicolumn{1}{c}{4.58} & \multicolumn{1}{c}{4.32} & \multicolumn{1}{c}{4.27} & \multicolumn{1}{c}{\bf 5.19} & \multicolumn{1}{c}{4.96} \\ \hline
		& \multicolumn{1}{c}{\bf GV(ResNet50)} & \multicolumn{1}{c}{4.57} & \multicolumn{1}{c}{7.91} &           \multicolumn{1}{c}{4.85} & \multicolumn{1}{c}{4.51} & \multicolumn{1}{c}{4.45} & \multicolumn{1}{c}{5.50} & \multicolumn{1}{c}{5.28} \\	
		&          {\bf GV(HRNet)}     &        {\bf 4.33}              &     7.41                 &        {\bf 4.51}              &      {\bf4.24}              &     {\bf 4.18}                 &     {\bf 5.19}                 &     {\bf 4.93}              \\ \hline
		\multirow{7}{*}{\begin{tabular}[c]{@{}c@{}}$FR_{10\%}(\%)$\\ (Lower is better)\end{tabular}} &        ESR\cite{cao2014face}                                    &           35.24           &        90.18              &   42.04                   &           30.80           &         38.84             &   47.28 & 41.40                    \\
		&          SDM\cite{xiong2013supervised}           &        29.40              &      84.36                &         33.44             &     26.22                 &     27.67                 &           41.85           &         35.32              \\
		&          CFSS\cite{zhu2015face}           &        20.56              &  66.26                    &         23.25             &       17.34             &  21.84                    &         32.88             &        23.67               \\
		&         DVLN\cite{wu2017leveraging}             &        10.84              &    46.93                  &       11.15               &           7.31           &    11.65                  &       16.30               &        13.71               \\ 
		&      LAB\cite{wu2018look} & 7.56  &  28.83  &      6.37         &      6.73             &       7.77               &   13.72                   &     10.74                  \\ 
		&         Wing\cite{feng2018wing}             &   6.00                   &     22.70                 &        4.78              &      4.30                &      7.77                &       12.50               &        7.76               \\ 
		&      AWing\cite{wang2019adaptive}             &    {\bf 2.84}                  &  {\bf 13.50}              &   {\bf 2.23}                   &     {\bf 2.58}                &    {\bf 2.91}                  &       {\bf 5.98}               &      {\bf 3.75}                \\ \hline
		&    {\bf GV(ResNet50)}     &    4.44             &    19.94                  &        3.18              &      3.72                &     3.88                 &    8.83                  &  6.47                     \\ 
		&    {\bf GV(HRNet)}        &    3.52           &     16.26                 &       2.55               &      3.30                &      3.40               &   6.79                   &   5.05                    \\ \hline
		\multirow{7}{*}{\begin{tabular}[c]{@{}c@{}}$AUC_{10\%}$\\ (Higher is better)\end{tabular}} &          ESR\cite{cao2014face}            &      0.2774& 0.0177& 0.1981 &0.2953& 0.2485& 0.1946 &0.2204                            \\
		&        SDM\cite{xiong2013supervised}              &       0.3002& 0.0226& 0.2293& 0.3237& 0.3125& 0.2060& 0.2398                        \\
		&         CFSS\cite{zhu2015face}            &     0.3659& 0.0632& 0.3157 &0.3854& 0.3691 &0.2688& 0.3037        \\
		&          DVLN\cite{wu2017leveraging}         &      0.4551& 0.1474& 0.3889& 0.4743& 0.4494& 0.3794& 0.3973                \\
		&         LAB\cite{wu2018look}             &     0.5323& 0.2345& 0.4951& 0.5433& 0.5394 &0.4490& 0.4630                 \\
		&         Wing\cite{feng2018wing}             &       0.5504& 0.3100& 0.4959& 0.5408& 0.5582& 0.4885& 0.4918                 \\
		&         AWing\cite{wang2019adaptive}     &    0.5719 &0.3120& 0.5149& 0.5777& 0.5715 &0.5022& 0.5120   \\ \hline
		& {\bf GV(ResNet50)}    &        0.5568              &     0.2769                         &      0.5321           &    0.5673                  &       0.5640               &      0.4782                &0.4959 \\
		&  {\bf GV(HRNet)}     &      {\bf 0.5775}         &      {\bf 0.3166}                &      {\bf 0.5636}                &      {\bf 0.5863}                &     {\bf 0.5881}                 &     {\bf 0.5035}     &     {\bf 0.5242}                  \\ \hline
	\end{tabular}
\end{table}
{\bf Evaluation on COFW}.
Experiments on COFW is displayed in Table~\ref{table:cofw}. 
Our baseline model GV(ResNet50) provides a better result than the previously best work AWing \cite{wang2019adaptive} with 4.92\% for IPN. 
GV(HRNet) achieves newly state-of-the-art performance with 4.85\% and reduces FR from 0.99\% to 0.39\%. 

{\bf Evaluation on WFLW}.
It is one of the most challenging benchmarks. 
We comprehensively compare the results of NME, FR@0.1, and AUC@0.1 with the previous works on the whole test set as well as all subsets in Table~\ref{table:wflw}. 
Our method achieves evident improvements of NME and AUC in almost all subsets.
Notably, ION on the whole test set decreases to 4.33\%.
We comprehensively compare our method with AWing in the supplement material.

{\bf Evaluation on JD-landmark}.
The results of JD-landmark are exhibited on Table~\ref{table:JD-landmark}. 
GV(HRNet) achieves comparable results (1.35\% for NME with FR 0.10\%) to the champion (1.31\% for NME with FR 0.10\%) with a single model excluding extra competition tricks. 
It is noteworthy that the champion \cite{liu2019grand} employed AutoML \cite{he2018amc} on architecture search, ensemble model, and a well-designed data augmentation scheme etc. 
The results further prove the briefness and effectiveness of our method for facial landmark detection.

\begin{minipage}{\textwidth}
	\begin{minipage}{0.5\textwidth}
		\begin{minipage}{1.0\textwidth}
			\tiny
			\renewcommand\arraystretch{1.4}
			\makeatletter\def\@captype{table}\makeatother
			\caption{The leadboard of JD-landmark}
			\label{table:JD-landmark}
			\begin{tabular}{cccc}
				\hline
				Ranking & Team          & AUC(@0.08)(\%)  & NME(\%) \\ \hline
				No.1    & Baidu VIS     & {\bf 84.01}                  & {\bf 1.31}    \\
				No.2    & USTC-NELSLIP  & 82.68                             & 1.41    \\
				No.3    & VIC iron man  & 82.22                             & 1.42    \\
				No.4    & CIGIT-ALIBABA & 81.53                              & 1.50    \\
				No.5    & Smiles        & 81.28                              & 1.50    \\ 
				Baseline    & -         & 73.32                             & 2.16    \\ \hline
				{\bf GV(HRNet)}  & -   & {\bf 83.34}                   & {\bf 1.35}  \\ \hline
			\end{tabular}
		\end{minipage}
		\vfill
		
		\begin{minipage}[!t]{1.0\textwidth}
			\renewcommand\arraystretch{1.4}
			\scriptsize
			\centering
			\makeatletter\def\@captype{table}\makeatother
			\caption{Ablation study on our method. 
				BBS means Beyond Box Strategy.
			}
			\begin{tabular}{ccc|ccc}
				\hline
				\multicolumn{1}{c}{\multirow{2}{*}{Backbone}} & \multicolumn{2}{c|}{\begin{tabular}[c]{@{}c@{}}Original face\\ bounding box\end{tabular}} & \multicolumn{3}{c}{\begin{tabular}[c]{@{}c@{}}Shrunk face\\ bounding box\end{tabular}} \\ \cline{2-6} 
				\multicolumn{1}{c}{}                          & \multicolumn{1}{c}{BPM}                    & \multicolumn{1}{c|}{NME(\%)}                    & \multicolumn{1}{c}{BPM}           & \multicolumn{1}{c}{BBS}           & NME(\%)        \\ \hline
				\multirow{2}{*}{ResNet50}   & -    &     3.42             &       $\checkmark$             &       -                             &    4.57          \\
				&      $\checkmark$                               &       {\bf 3.14}                                      &       $\checkmark$                              &      $\checkmark$                              &  {\bf 4.10}            \\ \hline
			\end{tabular}
			\label{table:compare model}
		\end{minipage}
	\end{minipage}
	\hfill
	\makeatletter\def\@captype{figure}\makeatother
	\begin{minipage}[!t]{0.5\textwidth}
		\centering
		\includegraphics[scale=0.4]{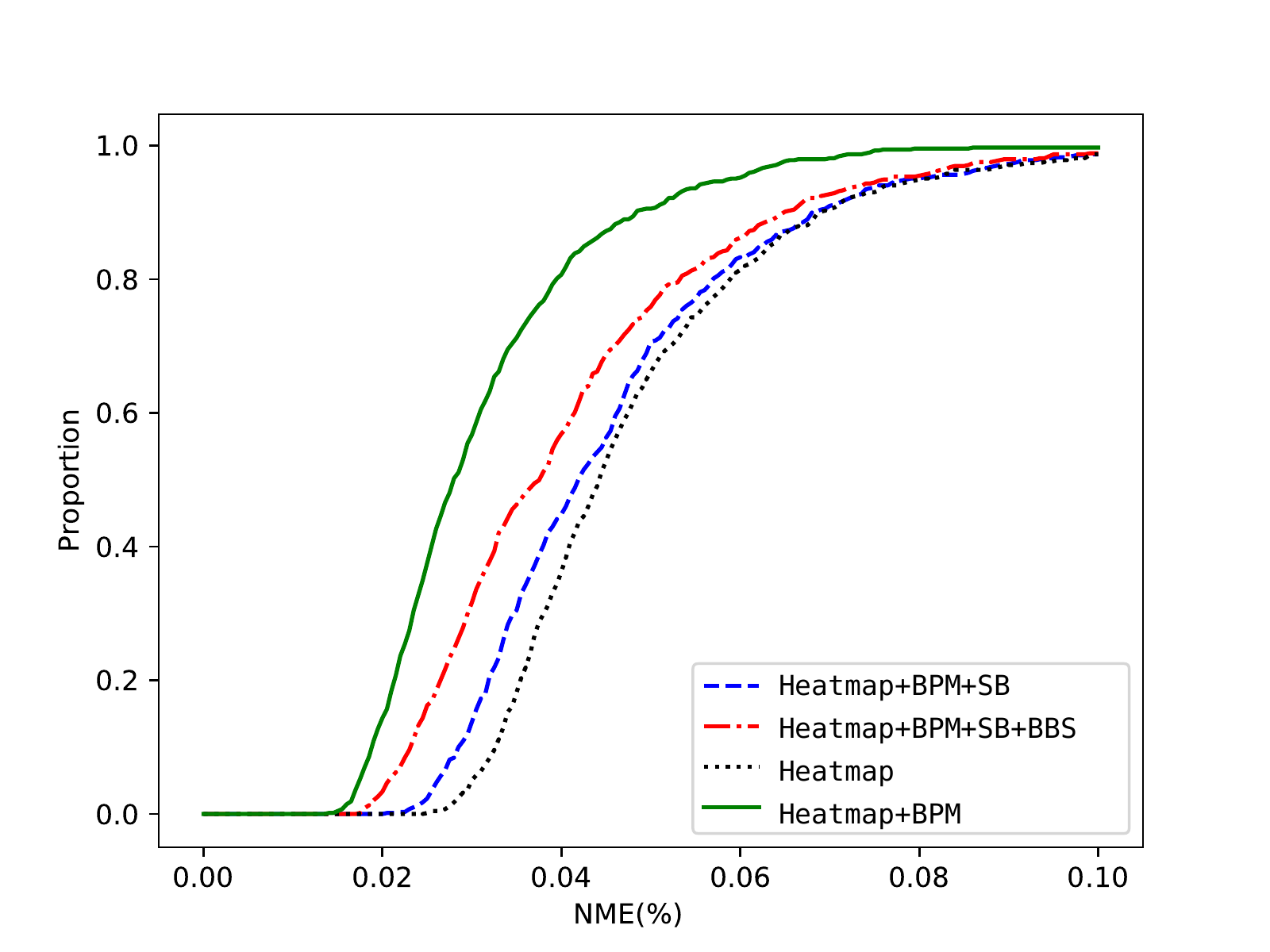}
		\caption{CED curves of models in Table~\ref{table:compare model}.
			SB means Shrinking the bounding box by 10\%.}
		\label{fig:ablation}
	\end{minipage}
\end{minipage}

\subsection{Ablation study}\label{section:ablation}
To demonstrate the effectiveness of Band Pooling Module and Beyond Box strategy, ablation study is implemented on 300W.
We choose ResNet50 as the backbone and ION as the test metric.

{\bf Effectiveness of BPM}.
We compare the performance with and without the BPM, that is, our proposed method and the heatmap based method. 
The result of each model is shown in Table~\ref{table:compare model}, and the BPM greatly improves the model performance from 3.42\% to 3.14\%.
Moreover, we provide CED curves to evaluate the accuracy
difference in Fig.~\ref{fig:ablation}.

\textbf{Choice of the bandwidth}. 
Experiments show that the choice of bandwidth $l$ has a great impact on the performance of the model in Table~\ref{table:bandwidth}.
It is obvious that oversize or undersize bandwidth does harm to the model performance (3.94\% when $l$=1 and 3.38\% when $l$=7). 
By aggregating different bandwidth, we get better localization results. 
Specifically, we obtain the best ION 3.14\% when fusing the vectors from bandwidth $l$=3 and $l$=5.

\textbf{Choice of $\theta$ in Eq. 2}.
We conduct a series of experiments on 300W with $\theta$ range from 0 to 4.
Note the backbone is ResNet50.
Experiments show that the appropriate $\theta$ gets better results in Table.~\ref{table:theta}

{\bf Improvements from the Beyond Box Strategy}. 
In Sec.~\ref{section:beyond}, we introduce a strategy to predict the landmarks out of the face bounding box.
Now a plain experiments is conducted to illustrate the effectiveness of it. 
We shrink the ground truth bounding box of 300W fullset by 10\% in each dimension to make some points fall out of the image. 
In this case, the prediction results turn worse (ION raised from 3.14\% to 4.57\%, shown in Table~\ref{table:compare model}) since the certain parts of face are invisible. 
We compare different choices for $\tau$ and $\rho$, as illustrated in Table~\ref{table:Beyod the Box}. When we pick $\tau$=2.0 and $\rho$=2.0, ION decreases from 4.57\% to 4.10\%.

\begin{minipage}{\textwidth}
	\begin{minipage}{0.5\textwidth}
	\makeatletter\def\@captype{table}\makeatother
	\begin{minipage}[!t]{0.9\textwidth}
	\scriptsize
	\renewcommand\arraystretch{1.2}
	\centering
	\caption{Different choices for $\tau$ and $\sigma$ in Beyond Box Strategy.}
	\setlength{\tabcolsep}{5pt}
	\begin{tabular}{cc|ccccc}
		\hline
		& $\sigma$            & \multirow{2}{*}{1.0}& \multirow{2}{*}{1.5}& \multirow{2}{*}{2.0} & \multirow{2}{*}{2.5} & \multirow{2}{*}{3.0}  \\
		$\tau$                        &                      &                      &                      &                      &            &          \\ \hline
		\multicolumn{2}{c|}{1.0}                       &        4.42         &    4.35                  &        4.29              &    4.26         &  4.24      \\ \hline
		\multicolumn{2}{c|}{1.5}               &        4.33          &   4.23                   &        4.15              &    4.12         &  4.14      \\ \hline
		\multicolumn{2}{c|}{2.0}                 &        4.28         &    4.17                  &        {\bf 4.10}              &    4.12         &  4.22      \\ \hline
		\multicolumn{2}{c|}{2.5}                &        4.26          &   4.15                   &        4.13              &    4.23          & 4.43      \\ \hline
	\end{tabular}
	\label{table:Beyod the Box}
	\end{minipage}
		\vfill
		
		\makeatletter\def\@captype{table}\makeatother
		\begin{minipage}[!t]{0.9\textwidth}
		\renewcommand\arraystretch{1.4}
		\scriptsize
		\centering
		\makeatletter\def\@captype{table}\makeatother
		\setlength{\tabcolsep}{4pt}
		\caption{Evaluation on different values of $\theta$.}
		\begin{tabular}{cccccc}
		\hline
		$\theta$ & 0 & 1 & 2 & 3 & 4 \\ \hline
		ION(\%)  & 3.32  & 3.23  & 3.21  &  \bf{3.14} & 3.25  \\ \hline
		\end{tabular}
		\label{table:theta}
		\end{minipage}
	\end{minipage}
	\hfill
	\makeatletter\def\@captype{table}\makeatother
	\begin{minipage}[!t]{0.5\textwidth}
	\renewcommand\arraystretch{1.4}
	\scriptsize
	\centering
	\makeatletter\def\@captype{table}\makeatother
	\setlength{\tabcolsep}{5pt}
	\caption{Comparison among different bandwidth. 
		Diagonal elements indicate that vector generated by single bandwidth BPM, as bandwidth $l$=1 or $l$=3.
		The others attained by aggregating different bandwidth where $\alpha$=0.5 in Eq.~\ref{equ:alpha}.}
	
	\begin{tabular}{c|cccc}
		\hline
		bandwidth 	& $l$=1 & $l$=3 & $l$=5&$l$=7 \\ \hline
		$l$=1 & 3.94  & 3.67  & 3.61 &  3.74  \\ \hline
		$l$=3 & -  & 3.31     & {\bf 3.14} &  3.22  \\ \hline
		$l$=5 & -  & -        &  3.21&  3.28   \\ \hline
		$l$=7 & -  & -        &  -   &  3.38    \\ \hline
	\end{tabular}
	\label{table:bandwidth}
	\end{minipage}
\end{minipage}

\subsection{Efficiency Analysis}\label{sec:analysis}
The efficiency of our method is examined in both convergence rate and time cost as follows.

\begin{figure}
	\centering
	\includegraphics[width=7.0cm]{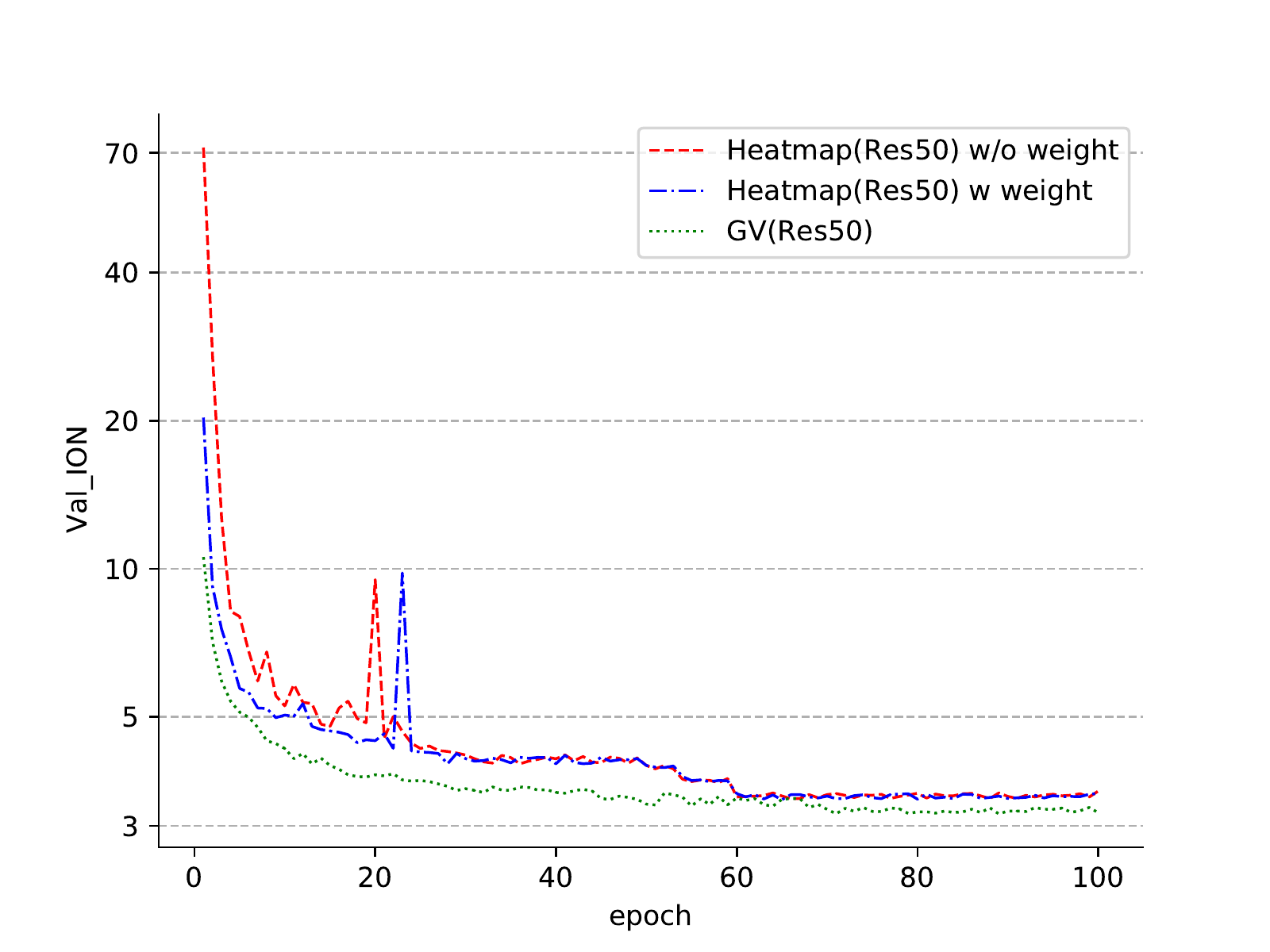}
	\caption{{\bf The training convergence plots of different method.}
		Note the logarithm of the y-axis here.
	}
	\label{fig:convergence}
\end{figure}

{\bf The comparision of convergence rate}.
We argue that the vector label speeds up model convergence and will experimentally demonstrate that.
In addition, a solution for imbalance has been proposed in AWing \cite{wang2019adaptive}.
By using a weight map, AWing applied different weights on hard examples(1.0) and easy examples(0.1) to alleviate the imbalance of heatmap.
Experiments on 300W show the impact of weight map and vector label on convergence.
We use the same learning rate and batch size for a fair comparison.
The metric is ION.
The results show both weight map and vector label facilitate the convergence and our method is better.

{\bf Time cost analysis} of Heatmap based method and Gaussian Vector. 
In embedded systems, the network runs on neural net accelerators (e.g. TPU) and post-processing on CPU (e.g. ARM).
The former accelerates common ops (Conv/Pool excluding argmax) significantly.
So BPM increases little computation complexity and time.
However, post-processing is the bottleneck since CPU has to process many other tasks synchronously.
For the limitation of the practical system (MT6735 with a NN accelerator), we use MobileNetV3 as the backbone to test the time cost of heatmap based method and ours.
As Table.~\ref{table:time_cost}, our method saves 26\% time of the whole process in this case.

\begin{table}
	\centering
	\renewcommand\arraystretch{1.2}
	\caption{{\bf Comparison of the time cost.}
	}
	\begin{tabular}{cccccc}
		\hline
		method & \begin{tabular}[c]{@{}c@{}}output\\tensor size\end{tabular} & \begin{tabular}[c]{@{}c@{}}net-forward\\(ms)\end{tabular} & \begin{tabular}[c]{@{}c@{}}post-processing\\(ms)\end{tabular} & \begin{tabular}[c]{@{}c@{}}transmission\\(ms)\end{tabular} \\ \hline
		Heatmap  & (68, 64, 64)  & 6.129  & 2.039  &  0.311  \\ \hline
		Gaussian Vector	& (68, 64, 2)  & 6.158  & 0.062  & 0.048 \\ \hline
	\end{tabular}
	\label{table:time_cost}
\end{table}


\section{Conclusion}

In this paper, we present a simple yet effective solution for facial landmark detection.
We provide novel vector label for supervision, Band Pooling Module to convert output heatmap into vectors, as well as Beyond Box Strategy for predicting landmarks out of the input.
By evaluating on different benchmarks, we demonstrate the efficiency of our proposed method.
Also, our method and its components are validated via comprehensive ablation study.



\bibliographystyle{splncs}
\bibliography{refer}

\begin{thebibliography}{10}

\bibitem{sun2014deep}
Sun, Y., Wang, X., Tang, X.:
\newblock Deep learning face representation from predicting 10,000 classes.
\newblock In: Proceedings of the IEEE conference on computer vision and pattern
  recognition. (2014)  1891--1898

\bibitem{liu2017sphereface}
Liu, W., Wen, Y., Yu, Z., Li, M., Raj, B., Song, L.:
\newblock Sphereface: Deep hypersphere embedding for face recognition.
\newblock In: Proceedings of the IEEE conference on computer vision and pattern
  recognition. (2017)  212--220

\bibitem{deng2019arcface}
Deng, J., Guo, J., Xue, N., Zafeiriou, S.:
\newblock Arcface: Additive angular margin loss for deep face recognition.
\newblock In: Proceedings of the IEEE Conference on Computer Vision and Pattern
  Recognition. (2019)  4690--4699

\bibitem{thies2016face2face}
Thies, J., Zollhofer, M., Stamminger, M., Theobalt, C., Nie{\ss}ner, M.:
\newblock Face2face: Real-time face capture and reenactment of rgb videos.
\newblock In: Proceedings of the IEEE conference on computer vision and pattern
  recognition. (2016)  2387--2395

\bibitem{dou2017end}
Dou, P., Shah, S.K., Kakadiaris, I.A.:
\newblock End-to-end 3d face reconstruction with deep neural networks.
\newblock In: Proceedings of the IEEE Conference on Computer Vision and Pattern
  Recognition. (2017)  5908--5917

\bibitem{roth2015unconstrained}
Roth, J., Tong, Y., Liu, X.:
\newblock Unconstrained 3d face reconstruction.
\newblock In: Proceedings of the IEEE conference on computer vision and pattern
  recognition. (2015)  2606--2615

\bibitem{feng2018joint}
Feng, Y., Wu, F., Shao, X., Wang, Y., Zhou, X.:
\newblock Joint 3d face reconstruction and dense alignment with position map
  regression network.
\newblock In: Proceedings of the European Conference on Computer Vision (ECCV).
  (2018)  534--551

\bibitem{zhu2017face}
Zhu, X., Liu, X., Lei, Z., Li, S.Z.:
\newblock Face alignment in full pose range: A 3d total solution.
\newblock IEEE transactions on pattern analysis and machine intelligence
  \textbf{41} (2017)  78--92

\bibitem{sun2013deep}
Sun, Y., Wang, X., Tang, X.:
\newblock Deep convolutional network cascade for facial point detection.
\newblock In: Proceedings of the IEEE conference on computer vision and pattern
  recognition. (2013)  3476--3483

\bibitem{zhang2014facial}
Zhang, Z., Luo, P., Loy, C.C., Tang, X.:
\newblock Facial landmark detection by deep multi-task learning.
\newblock In: European conference on computer vision, Springer (2014)  94--108

\bibitem{long2015fully}
Long, J., Shelhamer, E., Darrell, T.:
\newblock Fully convolutional networks for semantic segmentation.
\newblock In: Proceedings of the IEEE conference on computer vision and pattern
  recognition. (2015)  3431--3440

\bibitem{ronneberger2015u}
Ronneberger, O., Fischer, P., Brox, T.:
\newblock U-net: Convolutional networks for biomedical image segmentation.
\newblock In: International Conference on Medical image computing and
  computer-assisted intervention, Springer (2015)  234--241

\bibitem{zhao2020mobilefan}
Zhao, Y., Liu, Y., Shen, C., Gao, Y., Xiong, S.:
\newblock Mobilefan: Transferring deep hidden representation for face
  alignment.
\newblock Pattern Recognition \textbf{100} (2020)  107114

\bibitem{wang2019adaptive}
Wang, X., Bo, L., Fuxin, L.:
\newblock Adaptive wing loss for robust face alignment via heatmap regression.
\newblock In: Proceedings of the IEEE International Conference on Computer
  Vision. (2019)  6971--6981

\bibitem{kowalski2017deep}
Kowalski, M., Naruniec, J., Trzcinski, T.:
\newblock Deep alignment network: A convolutional neural network for robust
  face alignment.
\newblock In: Proceedings of the IEEE Conference on Computer Vision and Pattern
  Recognition Workshops. (2017)  88--97

\bibitem{wu2018look}
Wu, W., Qian, C., Yang, S., Wang, Q., Cai, Y., Zhou, Q.:
\newblock Look at boundary: A boundary-aware face alignment algorithm.
\newblock In: Proceedings of the IEEE conference on computer vision and pattern
  recognition. (2018)  2129--2138

\bibitem{newell2016stacked}
Newell, A., Yang, K., Deng, J.:
\newblock Stacked hourglass networks for human pose estimation.
\newblock In: European conference on computer vision, Springer (2016)  483--499

\bibitem{dong2018supervision}
Dong, X., Yu, S.I., Weng, X., Wei, S.E., Yang, Y., Sheikh, Y.:
\newblock Supervision-by-registration: An unsupervised approach to improve the
  precision of facial landmark detectors.
\newblock In: Proceedings of the IEEE Conference on Computer Vision and Pattern
  Recognition. (2018)  360--368

\bibitem{sagonas2013300}
Sagonas, C., Tzimiropoulos, G., Zafeiriou, S., Pantic, M.:
\newblock 300 faces in-the-wild challenge: The first facial landmark
  localization challenge.
\newblock In: Proceedings of the IEEE International Conference on Computer
  Vision Workshops. (2013)  397--403

\bibitem{burgos2013robust}
Burgos-Artizzu, X.P., Perona, P., Doll{\'a}r, P.:
\newblock Robust face landmark estimation under occlusion.
\newblock In: Proceedings of the IEEE international conference on computer
  vision. (2013)  1513--1520

\bibitem{liu2019grand}
Liu, Y., Shen, H., Si, Y., Wang, X., Zhu, X., Shi, H., Hong, Z., Guo, H., Guo,
  Z., Chen, Y.,  et~al.:
\newblock Grand challenge of 106-point facial landmark localization.
\newblock In: 2019 IEEE International Conference on Multimedia \& Expo
  Workshops (ICMEW), IEEE (2019)  613--616

\bibitem{cootes2001active}
Cootes, T.F., Edwards, G.J., Taylor, C.J.:
\newblock Active appearance models.
\newblock IEEE Transactions on pattern analysis and machine intelligence
  \textbf{23} (2001)  681--685

\bibitem{kahraman2007active}
Kahraman, F., Gokmen, M., Darkner, S., Larsen, R.:
\newblock An active illumination and appearance (aia) model for face alignment.
\newblock In: 2007 IEEE Conference on Computer Vision and Pattern Recognition,
  IEEE (2007)  1--7

\bibitem{matthews2004active}
Matthews, I., Baker, S.:
\newblock Active appearance models revisited.
\newblock International journal of computer vision \textbf{60} (2004)  135--164

\bibitem{milborrow2008locating}
Milborrow, S., Nicolls, F.:
\newblock Locating facial features with an extended active shape model.
\newblock In: European conference on computer vision, Springer (2008)  504--513

\bibitem{cristinacce2006feature}
Cristinacce, D., Cootes, T.F.:
\newblock Feature detection and tracking with constrained local models.
\newblock In: Bmvc. Volume~1., Citeseer (2006) ~3

\bibitem{cao2014face}
Cao, X., Wei, Y., Wen, F., Sun, J.:
\newblock Face alignment by explicit shape regression.
\newblock International Journal of Computer Vision \textbf{107} (2014)
  177--190

\bibitem{chen2014joint}
Chen, D., Ren, S., Wei, Y., Cao, X., Sun, J.:
\newblock Joint cascade face detection and alignment.
\newblock In: European conference on computer vision, Springer (2014)  109--122

\bibitem{xiong2015global}
Xiong, X., De~la Torre, F.:
\newblock Global supervised descent method.
\newblock In: Proceedings of the IEEE Conference on Computer Vision and Pattern
  Recognition. (2015)  2664--2673

\bibitem{trigeorgis2016mnemonic}
Trigeorgis, G., Snape, P., Nicolaou, M.A., Antonakos, E., Zafeiriou, S.:
\newblock Mnemonic descent method: A recurrent process applied for end-to-end
  face alignment.
\newblock In: Proceedings of the IEEE Conference on Computer Vision and Pattern
  Recognition. (2016)  4177--4187

\bibitem{feng2018wing}
Feng, Z.H., Kittler, J., Awais, M., Huber, P., Wu, X.J.:
\newblock Wing loss for robust facial landmark localisation with convolutional
  neural networks.
\newblock In: Proceedings of the IEEE Conference on Computer Vision and Pattern
  Recognition. (2018)  2235--2245

\bibitem{sun2019high}
Sun, K., Zhao, Y., Jiang, B., Cheng, T., Xiao, B., Liu, D., Mu, Y., Wang, X.,
  Liu, W., Wang, J.:
\newblock High-resolution representations for labeling pixels and regions.
\newblock arXiv preprint arXiv:1904.04514 (2019)

\bibitem{sun2019deep}
Sun, K., Xiao, B., Liu, D., Wang, J.:
\newblock Deep high-resolution representation learning for human pose
  estimation.
\newblock In: Proceedings of the IEEE Conference on Computer Vision and Pattern
  Recognition. (2019)  5693--5703

\bibitem{liu2018intriguing}
Liu, R., Lehman, J., Molino, P., Such, F.P., Frank, E., Sergeev, A., Yosinski,
  J.:
\newblock An intriguing failing of convolutional neural networks and the
  coordconv solution.
\newblock In: Advances in Neural Information Processing Systems. (2018)
  9605--9616

\bibitem{sun2018integral}
Sun, X., Xiao, B., Wei, F., Liang, S., Wei, Y.:
\newblock Integral human pose regression.
\newblock In: Proceedings of the European Conference on Computer Vision (ECCV).
  (2018)  529--545

\bibitem{nibali2018numerical}
Nibali, A., He, Z., Morgan, S., Prendergast, L.:
\newblock Numerical coordinate regression with convolutional neural networks.
\newblock arXiv preprint arXiv:1801.07372 (2018)

\bibitem{levine2016end}
Levine, S., Finn, C., Darrell, T., Abbeel, P.:
\newblock End-to-end training of deep visuomotor policies.
\newblock The Journal of Machine Learning Research \textbf{17} (2016)
  1334--1373

\bibitem{lin2017feature}
Lin, T.Y., Doll{\'a}r, P., Girshick, R., He, K., Hariharan, B., Belongie, S.:
\newblock Feature pyramid networks for object detection.
\newblock In: Proceedings of the IEEE conference on computer vision and pattern
  recognition. (2017)  2117--2125

\bibitem{le2012interactive}
Le, V., Brandt, J., Lin, Z., Bourdev, L., Huang, T.S.:
\newblock Interactive facial feature localization.
\newblock In: European conference on computer vision, Springer (2012)  679--692

\bibitem{belhumeur2013localizing}
Belhumeur, P.N., Jacobs, D.W., Kriegman, D.J., Kumar, N.:
\newblock Localizing parts of faces using a consensus of exemplars.
\newblock IEEE transactions on pattern analysis and machine intelligence
  \textbf{35} (2013)  2930--2940

\bibitem{zhu2012face}
Zhu, X., Ramanan, D.:
\newblock Face detection, pose estimation, and landmark localization in the
  wild.
\newblock In: 2012 IEEE conference on computer vision and pattern recognition,
  IEEE (2012)  2879--2886

\bibitem{kingmaadam}
Kingma, D.P., Ba, J.L.:
\newblock
\newblock (Adam: Amethod for stochastic optimization)

\bibitem{chen2015mxnet}
Chen, T., Li, M., Li, Y., Lin, M., Wang, N., Wang, M., Xiao, T., Xu, B., Zhang,
  C., Zhang, Z.:
\newblock Mxnet: A flexible and efficient machine learning library for
  heterogeneous distributed systems.
\newblock arXiv preprint arXiv:1512.01274 (2015)

\bibitem{ren2014face}
Ren, S., Cao, X., Wei, Y., Sun, J.:
\newblock Face alignment at 3000 fps via regressing local binary features.
\newblock In: Proceedings of the IEEE Conference on Computer Vision and Pattern
  Recognition. (2014)  1685--1692

\bibitem{xiao2016robust}
Xiao, S., Feng, J., Xing, J., Lai, H., Yan, S., Kassim, A.:
\newblock Robust facial landmark detection via recurrent attentive-refinement
  networks.
\newblock In: European conference on computer vision, Springer (2016)  57--72

\bibitem{valle2018deeply}
Valle, R., Buenaposada, J.M., Valdes, A., Baumela, L.:
\newblock A deeply-initialized coarse-to-fine ensemble of regression trees for
  face alignment.
\newblock In: Proceedings of the European Conference on Computer Vision (ECCV).
  (2018)  585--601

\bibitem{feng2017dynamic}
Feng, Z.H., Kittler, J., Christmas, W., Huber, P., Wu, X.J.:
\newblock Dynamic attention-controlled cascaded shape regression exploiting
  training data augmentation and fuzzy-set sample weighting.
\newblock In: Proceedings of the IEEE Conference on Computer Vision and Pattern
  Recognition. (2017)  2481--2490

\bibitem{kumar2018disentangling}
Kumar, A., Chellappa, R.:
\newblock Disentangling 3d pose in a dendritic cnn for unconstrained 2d face
  alignment.
\newblock In: Proceedings of the IEEE Conference on Computer Vision and Pattern
  Recognition. (2018)  430--439

\bibitem{he2016identity}
He, K., Zhang, X., Ren, S., Sun, J.:
\newblock Identity mappings in deep residual networks.
\newblock In: European conference on computer vision, Springer (2016)  630--645

\bibitem{xiong2013supervised}
Xiong, X., De~la Torre, F.:
\newblock Supervised descent method and its applications to face alignment.
\newblock In: Proceedings of the IEEE conference on computer vision and pattern
  recognition. (2013)  532--539

\bibitem{zhu2015face}
Zhu, S., Li, C., Change~Loy, C., Tang, X.:
\newblock Face alignment by coarse-to-fine shape searching.
\newblock In: Proceedings of the IEEE conference on computer vision and pattern
  recognition. (2015)  4998--5006

\bibitem{wu2017leveraging}
Wu, W., Yang, S.:
\newblock Leveraging intra and inter-dataset variations for robust face
  alignment.
\newblock In: Proceedings of the IEEE conference on computer vision and pattern
  recognition workshops. (2017)  150--159

\bibitem{he2018amc}
He, Y., Lin, J., Liu, Z., Wang, H., Li, L.J., Han, S.:
\newblock Amc: Automl for model compression and acceleration on mobile devices.
\newblock In: Proceedings of the European Conference on Computer Vision (ECCV).
  (2018)  784--800

\end{thebibliography}

\section{supplementary material}
\subsection{About FR@(10\%) in Table 3}
In Table 3, we achieve better NME and AUC than AWing[42], but worse FR.
As we presented in Sec 4.2, NME calculates the overall mean error of the test set, and AUC evaluates the samples with error lower than threshold.
But FR is the proportion of samples with error higher than threshold in the test set. 
The results show that, compared with AWing, our approach is better overall but worse in extreme hard samples. (e.g. heavy occlusion).
To achieve SOTA, AWing used boundary map and multi-stage supervision, resulting in increased network complexity.
As first introduced in the LAB [43], the boundary map provides better face structure information to deal with hard samples.
However, it is complicated to generate and not suitable for sparse landmarks (e.g. 5 points).
To prove our vector label works, we try to keep our network simple enough.
Even so, our method goes beyond AWing on the NME and AUC.

\subsection{Analysis}
\begin{figure}
	\centering
	\includegraphics[width=10.0cm]{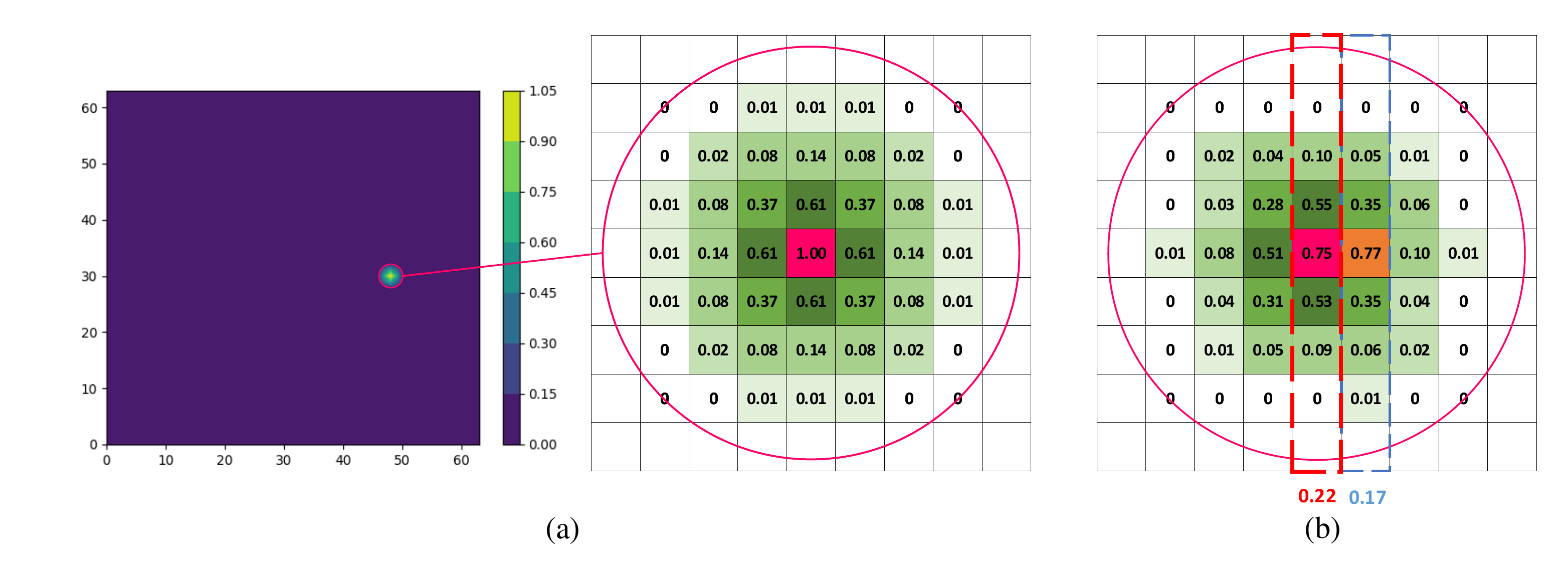}
	\caption{(a) The ground truth Gaussian heatmap. 
		(b) A tiny shift on the pixel with the highest response causes location error in predicted heatmap.
		The ground truth location (\textit{red} pixel with 0.75) is smaller than the one on the right (\textit{orange} pixel with 0.77), bringing about extra location error. 
		With $l$=1, however, the mean of corresponding two columns is 0.22 (\textit{red} dashed region) and 0.17 (\textit{blue} dashed region). 
		The vertical band pooling helps eliminate the error by average the pixels in the band.
	}
	\label{fig:amend}
\end{figure}

We can see that our proposed method surpasses the previous heatmap based methods by a large margin.
{\bf How do the vector supervision and BPM contribute to model performance?} Here we analyze the merits in three aspects.

Firstly, vector supervision alleivates the imbalance of foreground and background to a great extent.
For example, in the typical Gaussian heatmap with a shape of $64\times64$, standard deviation $\sigma$=2, the foreground pixels take up only 4\%.
The extreme ratio leads the foreground to a less important position.
In the early training stage, the optimizer makes great effort to push all pixels to zero values, which slows down the convergence.
Even after a period of training, the accumulation of small errors on the background still affects the training optimization, which is meaningless for landmark detection.
The proportion rises to 20\% when vector supervision is used.
Consequently, the whole training process becomes more efficient.

Secondly, in the heatmap based methods, only the pixels in the neighbor domain of the maximum one are used to get the final landmark coordinates.
Most of the spatial information is discarded even though we devote much energy to optimize it.
Worse yet, a tiny shift of the maximum pixel on the heatmap may increase location error.
Fig.~\ref{fig:amend}(a) shows heatmap supervision and Fig.~\ref{fig:amend}(b) shows maximum pixel shifting on the heatmap prediction incurs an evident error.
Overwhelming the heatmap based methods, the proposed Band Pooling Module comprehensively makes use of the spatial information.
It demonstrates that BPM stengthens the resistibility of the maximum pixel shifting and leads to more robust prediction.

\begin{figure}
	\centering
	\includegraphics[width=8.0cm]{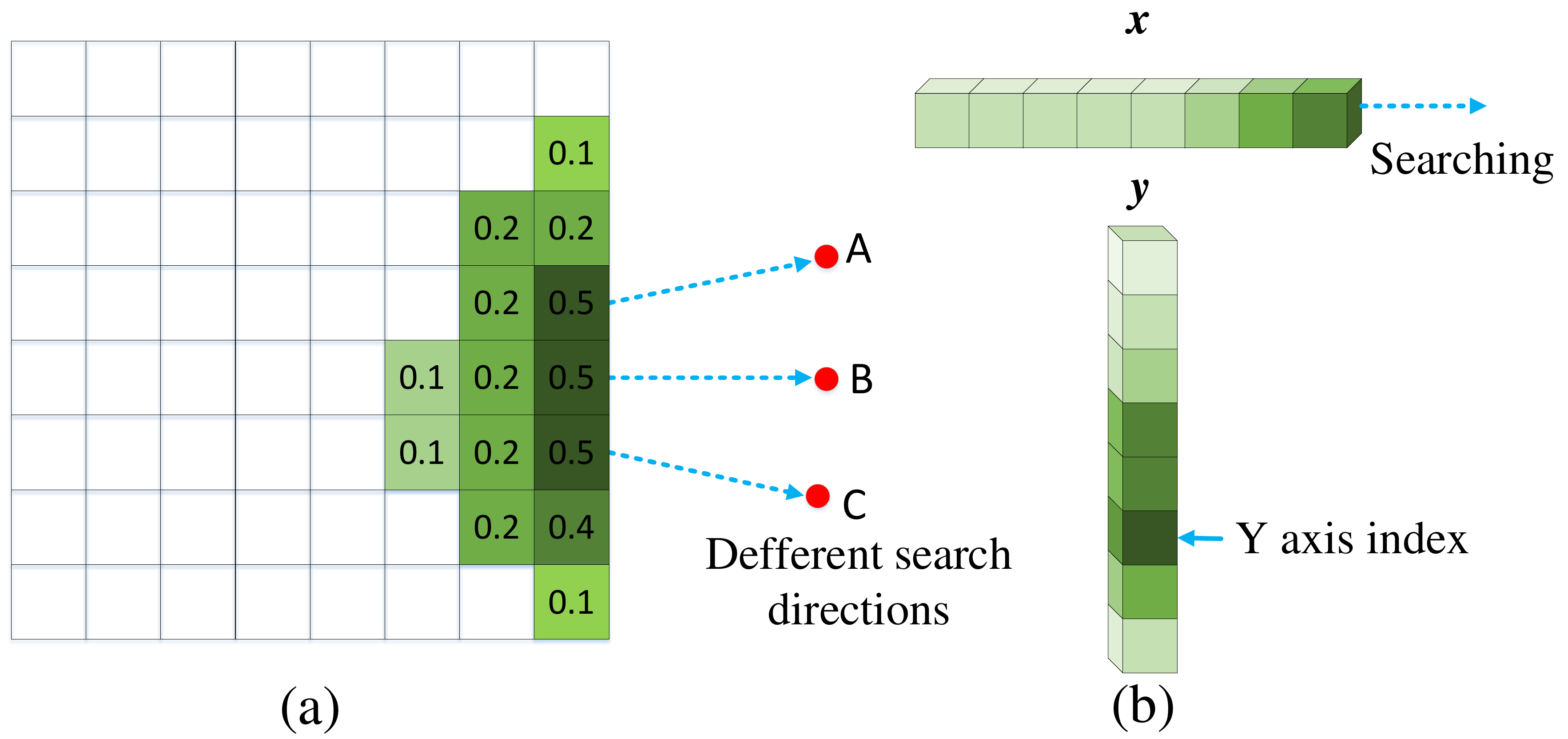}
	\caption{(a) The maximum pixel locates on the right edge of the heatmap. 
		Regression error caused by weak supervision usually brings about multiple maximum pixel or maximum pixel shifting. 
		Therefore, it is difficult to locate the correct starting point and search direction. 
		(b) Vector supervision decouples $X$ and $Y$ axes, as a result, we are able to locate the correct Y axis index from vector $\bf y$ and search along the right direction.
	}
	\label{fig:direction}
\end{figure}

Finally, Beyond Box Strategy helps our model look outside and predict the landmarks out of the bounding box.
With this strategy, we search the peak of quasi-Gaussian distribution towards a single direction.
However, it is too hard to assume an exact 2D distribution in a similar way for heatmap based methods, because we are usually not sure about the starting point and search direction.
As shown in Fig.~\ref{fig:direction}(a), multiple maximum pixels locate on the right edge of the heatmap.
As a result, we cannot determine which point to start with and which direction to search towards.
Since we convert the heamtp into a pair of vectors like Fig.~\ref{fig:direction}(b), the two axes are decorrelated naturally.
We can easily get the accurate starting point from the maximum of vector ${\bf y}$, and search on vector ${\bf x}$ along the right side.
Therefore, even if the landmarks fall outside, Beyond Box Strategy is able to settle this trouble in most cases.

\end{document}